\documentclass{article}

\usepackage{PRIMEarxiv}

\usepackage[utf8]{inputenc} 
\usepackage[T1]{fontenc}    
\usepackage{hyperref}       
\usepackage{url}            
\usepackage{booktabs}       
\usepackage{amsfonts}       
\usepackage{nicefrac}       
\usepackage{microtype}      
\usepackage{fancyhdr}       
\usepackage{graphicx}       
\usepackage{amsmath}
\usepackage{array}
\usepackage{float}
\graphicspath{{media/}}     

\title{
ArXiv-to-Model: A Practical Study of Scientific LM Training
}

\author{
Anuj Gupta \\
Independent Researcher \\
India \\
\texttt{anuj0456@gmail.com}
}

\begin{document}
\maketitle

\begin{abstract}
While frontier large language models demonstrate strong reasoning and 
mathematical capabilities, the practical process of training 
domain-specialized scientific language models from raw sources 
remains under-documented. In this work, we present a detailed case 
study of training a 1.36B-parameter scientific language model 
directly from raw arXiv LaTeX sources spanning mathematics, 
computer science, and theoretical physics.

We describe an end-to-end pipeline covering metadata filtering, 
archive validation, LaTeX extraction, text normalization, 
domain-aware tokenization, and dense transformer training 
under constrained compute (2$\times$A100 GPUs). Through 24 
experimental runs, we analyze training stability, scaling behavior, 
data yield losses, and infrastructure bottlenecks.

Our findings highlight how preprocessing decisions significantly 
affect usable token volume, how tokenization impacts symbolic 
stability, and how storage and I/O constraints can rival compute 
as limiting factors. We further analyze convergence dynamics 
and show stable training behavior in a data-rich regime 
(52B pretraining tokens).

Rather than proposing a novel architecture, this work 
provides an engineering-grounded, transparent account of 
training a small scientific language model from scratch. 
We hope these insights support researchers operating under 
moderate compute budgets who seek to build domain-specialized models.
\end{abstract}

\keywords{Scientific Language Models \and ArXiv \and Language Model Training \and Machine Learning}

\section{Introduction}

Recent advances in large language models have demonstrated 
impressive reasoning, mathematical, and scientific capabilities. 
However, the majority of publicly reported systems rely on 
large-scale curated corpora, proprietary data mixtures, 
or undisclosed preprocessing pipelines. In contrast, the 
practical process of constructing a scientific language model 
directly from raw open-access sources remains sparsely documented.

Training domain-specific models from raw scientific corpora 
introduces unique challenges. ArXiv distributions consist of 
heterogeneous LaTeX archives, multi-file project structures, 
custom macros, symbolic-heavy content, and inconsistent metadata. 
Seemingly minor preprocessing choices—such as language filtering 
heuristics or archive validation rules—can significantly impact 
final dataset yield and training stability.

In this paper, we document our experience training a 
1.36B-parameter scientific language model using raw arXiv sources. 
Our goal is not to introduce a new architecture nor to compete 
with large instruction-tuned systems, but rather to provide a 
transparent and reproducible account of:

\begin{itemize}
    \item End-to-end dataset construction from raw metadata and LaTeX archives
    \item Tokenization strategies for formula-dense scientific text
    \item Training dynamics under constrained GPU and storage budgets
    \item Empirical scaling behavior in a data-rich regime
\end{itemize}

Through 24 iterative runs, we analyze failure modes, 
optimization instabilities, preprocessing bottlenecks, 
and hardware utilization patterns. We compare small-data 
(20GB) and full-data (200GB) regimes to illustrate the 
impact of dataset scale on convergence behavior.

Our contributions are primarily empirical and engineering-focused:

\begin{enumerate}
    \item A reproducible pipeline for constructing a 
    scientific corpus from raw arXiv LaTeX sources.
    \item Quantitative analysis of dataset yield losses 
    caused by metadata filtering and extraction failures.
    \item Practical insights into tokenizer design for 
    mathematical and symbolic-heavy corpora.
    \item A detailed examination of training stability 
    across 24 experimental runs.
\end{enumerate}

We believe that transparent reporting of data engineering, 
infrastructure tradeoffs, and training dynamics is essential 
for advancing reproducible scientific language modeling—particularly 
for researchers operating under limited compute resources.

\section{Training Data Scaling Considerations}
The backbone of this study is a 80GB text corpus constructed from arXiv LaTeX sources. 
The dataset construction pipeline was divided into four major stages: (1) source extraction, (2) metadata filtering, (3) LaTeX normalization and cleaning, and (4) weighted mixture assembly.

\subsection{How much data (tokens) do you need?}

Following the Chinchilla scaling law, the optimal number of training 
tokens scales approximately linearly with model parameters:

\begin{equation}
T \approx 20 \times P
\end{equation}

where $T$ denotes the number of training tokens and $P$ the number of 
model parameters. This compute-optimal regime suggests that 
under-training (too few tokens) leads to underutilized model capacity, 
while over-training wastes compute without proportional gains.

\subsection{Converting Text Size to Tokens}

A commonly used empirical approximation for clean English text is:

\begin{quote}
$1$ billion tokens $\approx 3$--$4$~GB of processed text
\end{quote}

Using this approximation:

\begin{itemize}
    \item $40$B tokens $\approx 120$--$160$~GB
    \item $140$B tokens $\approx 420$--$560$~GB
\end{itemize}

Note that scientific LaTeX corpora often exhibit slightly different 
compression characteristics due to symbolic density and equation-heavy 
content.

\subsection{Practical Data Scale Regimes}

From empirical training observations and scaling literature, 
we categorize dataset sizes as follows:

\begin{itemize}
    \item $<$10~GB $\rightarrow$ Too small for pretraining; suitable primarily for fine-tuning.
    \item 20--50~GB $\rightarrow$ Borderline regime; requires many epochs and risks overfitting.
    \item 100--300~GB $\rightarrow$ Suitable for training $\sim$1.5B parameter models in a compute-efficient regime.
\end{itemize}

\subsection{Metadata Filtering and Normalization}

Raw arXiv distributions present several structural and quality challenges. Submissions often contain multiple versions, short-form letters, withdrawn manuscripts, and non-English text. Additionally, many papers are organized as multi-file LaTeX projects linked via \texttt{\textbackslash input} or \texttt{\textbackslash include} directives, frequently relying on custom macros or external style files. These factors introduce significant noise 
into large-scale scientific language model training.

To ensure corpus consistency and quality, we implemented a multi-stage filtering pipeline with the following constraints:

\begin{itemize}
    \item \textbf{Subject Focus:} Restricted to \texttt{math}, \texttt{cs}, \texttt{hep-th}, 
    \texttt{hep-ph}, \texttt{quant-ph}, \texttt{stat.ML}, and \texttt{stat.TH} categories to preserve formal scientific reasoning structure.
    
    \item \textbf{Temporal Filter:} Only papers published after the year 2000 were retained, reducing legacy formatting inconsistencies and outdated stylistic conventions.
    
    \item \textbf{Withdrawal Removal:} Submissions marked as withdrawn in metadata comments were excluded to prevent inclusion of invalidated or retracted work.
    
    \item \textbf{Volume Filter:} Documents with body text under 2,000 characters were removed to ensure exposure to complete scientific arguments rather than fragments or short notes.
    
    \item \textbf{Language Detection:} Automated language identification was applied to the combined title, abstract, and cleaned body text. Due to dense mathematical notation, occasional false negatives were observed; however, this step effectively removed clearly non-English submissions.
    
    \item \textbf{Archive Validation:} Each downloaded source archive was verified for structural integrity before extraction to prevent malformed or corrupted tar files from entering the pipeline.
\end{itemize}

All available \texttt{.tex} sources within validated archives were extracted and concatenated. 
The cleaning stage removed figures, references, formatting commands, and non-semantic LaTeX artifacts while preserving mathematical expressions and structural environments essential for scientific reasoning.

To mitigate redundancy, we applied both exact deduplication via content hashing and near-duplicate detection using similarity-based thresholding. This reduces repeated content arising from versioned submissions or minimally modified re-uploads while retaining substantive revisions.

Despite these efforts, LaTeX extraction remains a primary source of unavoidable data loss due to malformed sources or unconventional project layouts.

\subsection{Sampling Weights and Data Mixture}

To balance expertise and stylistic diversity, we implemented a weighted sampling strategy during mixture assembly. High-quality ``Gold'' scientific documents were upsampled to increase exposure to precise terminology, formal proofs, and domain-specific structure. Simultaneously, broader domain papers were retained at lower sampling weights to prevent overfitting to narrow stylistic patterns while preserving generalization capacity.

This weighted mixture enables the model to achieve strong technical competence without sacrificing robustness across adjacent scientific subdomains.

\begin{table} [H]
\centering
\begin{tabular}{|l|c|l|c|>{\raggedright\arraybackslash}p{4cm}|}
\hline
Source & Volume & Stage & Weight & Relevance \\ 
\hline
\hline
Cleaned arXiv (LaTeX) & 80GB & Pretraining & 2.0X & Core scientific reasoning, formal proofs, mathematical structure \\ 
OpenWebMath & 50GB & Pretraining & 1.0X & Informal mathematical intuition and diverse problem formulations \\ 
StackExchange (STEM) & 10GB & Post-training & 1.0X & Logical question-answer reasoning and structured explanations \\ 
MathInstruct & 100MB & Post-training & 1.0X & Supervised problem-solving and step-by-step reasoning \\ 
UltraChat & 1.2GB & Post-training & 1.0X & Conversational alignment and dialogue formatting \\ 
\hline
\end{tabular}
\caption{Training Data Sources, Stages, and Sampling Weights}
\label{tab:data_mixture}
\end{table}

\paragraph{Pretraining vs Post-Training Strategy.}
The base model was pretrained exclusively on LaTeX-formatted scientific corpora, primarily cleaned arXiv sources. This ensures that the model internalizes formal mathematical structure, symbolic reasoning, theorem-proof patterns, and domain-specific terminology without contamination from conversational noise.

External datasets (e.g., StackExchange, MathInstruct, UltraChat) were introduced only during post-training. Their purpose is not to teach scientific knowledge, but to align the model to instruction-following behavior, conversational formatting, and structured reasoning outputs. 

This separation allows the model to first develop deep scientific competence 
before being aligned for interactive usage, minimizing catastrophic interference 
between formal reasoning and conversational fluency.

\section{Tokenization}

Scientific and mathematical corpora differ fundamentally from general natural language. They contain dense symbolic expressions, structured equations, operator heavy sequences, and domain-specific LaTeX environments. Generic tokenization schemes optimized for web text or conversational data tend to fragment mathematical symbols, over-segment operators, and distort structural boundaries within equations.

Such fragmentation negatively impacts compression efficiency, increases sequence length, and weakens the model’s ability to learn stable representations of formal reasoning patterns. For scientific language modeling, tokenization is therefore not a trivial preprocessing step but a core architectural design decision.

\subsection{Design Objectives}

The tokenization pipeline was developed with the following objectives:

\begin{itemize}
    \item Preserve mathematical expressions and LaTeX structural environments.
    \item Reduce unnecessary fragmentation of symbols, operators, and formula blocks.
    \item Improve token compression efficiency for equation-heavy documents.
    \item Maintain representational consistency across scientific subdomains.
\end{itemize}

\subsection{Methodology}

We conducted exploratory experiments training custom BPE and 
SentencePiece tokenizers on curated subsets of the scientific corpus. 
These experiments aimed to:

\begin{itemize}
    \item Preserve common LaTeX commands and operators
    \item Reduce fragmentation of symbolic expressions
    \item Improve compression efficiency in equation-heavy documents
\end{itemize}

Preliminary trials revealed that tokenizer sampling strategy and 
vocabulary size significantly influenced symbolic segmentation patterns. 
However, integrating a newly trained tokenizer introduced additional 
complexity in embedding alignment and model initialization stability.

\subsection{Final Tokenizer Selection}

For the final KiteFish-A1-1.5B model, we adopted a LLaMA-compatible 
SentencePiece tokenizer with a vocabulary size of approximately 
102,400 tokens.

This decision was motivated by:

\begin{itemize}
    \item Architectural compatibility with the LLaMA transformer design
    \item Stable embedding initialization
    \item Reduced risk of token-ID misalignment
    \item Consistent convergence across experimental runs
\end{itemize}

Although domain-trained tokenizers remain a promising direction, 
the LLaMA tokenizer demonstrated sufficient robustness for 
scientific text modeling under current compute constraints.

\subsection{Tokenization Output Statistics}

After tokenization of approximately 200~GB of curated scientific data, 
the resulting corpus yielded:

\begin{itemize}
    \item 52.18B tokens for scientific pretraining
    \item 5B tokens for post-training and alignment data
\end{itemize}

The token density reflects the symbolic compression characteristics 
of scientific LaTeX, where formula-heavy documents produce 
distinct segmentation behavior relative to general web corpora.

\subsection{Evaluation Metrics}

Tokenizer evaluation during exploratory experiments considered:

\begin{itemize}
    \item Average tokens per document (compression efficiency)
    \item Symbol fragmentation patterns
    \item Early-stage training stability
\end{itemize}

Future work may systematically compare domain-trained tokenizers 
against general-purpose tokenizers for symbolic efficiency 
and long-context reasoning.

\section{Model Architecture}

The model is implemented as a dense, decoder-only transformer 
following the LLaMA architectural framework. The final configuration 
contains approximately \textbf{1.36 billion parameters}, including 
untied input and output embeddings.

\subsection{Architectural Specification}

The architectural configuration is summarized below:

\begin{itemize}
    \item Hidden dimension ($d_{\text{model}}$): 2048
    \item Transformer layers: 24
    \item Attention heads: 16
    \item Key–value heads: 16 (standard multi-head attention)
    \item Feed-forward dimension: 5504
    \item Vocabulary size: 102{,}400
    \item Positional encoding: Rotary Position Embeddings (RoPE, $\theta=10{,}000$)
    \item Maximum context length: 4096 tokens
    \item Activation function: SiLU
    \item Normalization: RMSNorm ($\epsilon=10^{-6}$)
    \item Precision: bfloat16
\end{itemize}

Unlike embedding-tied configurations, input and output token embeddings 
are maintained as separate matrices, increasing representational flexibility 
at the cost of additional parameters. The resulting model contains 
approximately 1.36B trainable parameters.

\subsection{Design Rationale}

A dense transformer architecture was selected over sparse or 
Mixture-of-Experts (MoE) alternatives for the following reasons:

\begin{itemize}
    \item \textbf{Training Stability:} Dense models exhibit more predictable 
    convergence behavior under moderate-scale compute.
    
    \item \textbf{Deterministic Compute Per Token:} Unlike MoE routing, 
    each token activates all parameters within a layer, simplifying 
    optimization dynamics and distributed training.
    
    \item \textbf{Efficient Multi-GPU Scaling:} Dense architectures 
    reduce cross-device communication overhead relative to expert 
    routing strategies.
    
    \item \textbf{Domain Specialization:} Given the high-quality, 
    domain-focused scientific corpus, parameter efficiency was prioritized 
    over sparse capacity scaling.
\end{itemize}

\subsection{Compute Budget and Infrastructure}

Training was conducted on:

\begin{itemize}
    \item 2$\times$ NVIDIA A100 GPUs (80GB memory each)
    \item Distributed data-parallel setup
    \item High-throughput storage-backed dataset streaming
\end{itemize}

Projected compute usage for the primary training phase 
was approximately 5{,}000--8{,}000 GPU-hours.

Efficiency was maximized using:

\begin{itemize}
    \item bfloat16 mixed-precision training
    \item Activation checkpointing
    \item Optimized data loading pipelines
\end{itemize}

\subsection{Batch Size and Parallelization Strategy}

To ensure stable optimization and efficient hardware utilization, 
training employed data parallelism combined with gradient accumulation.

\begin{itemize}
    \item Micro-batch size per GPU: 1--2 sequences
    \item Effective global batch size: 512--2{,}048 sequences
    \item Gradient accumulation: tuned dynamically to match memory constraints
\end{itemize}

Fully Sharded Data Parallel (FSDP) or ZeRO-style optimization 
was used to enable scalable memory partitioning across GPUs 
while maintaining numerical stability.

\subsection{Dense vs Sparse Capacity Tradeoffs}

While Mixture-of-Experts (MoE) architectures increase total 
parameter capacity without proportional inference cost, they 
introduce routing complexity, expert imbalance, and higher 
distributed communication overhead.

Given the available compute budget (2$\times$A100 GPUs) and 
the objective of scientific specialization rather than 
parameter-scale maximization, a dense transformer was selected 
for stability, deterministic compute per token, and efficient 
multi-GPU scaling.

\section{Training Setup}

The training pipeline was designed to balance scientific rigor, 
hardware constraints, and optimization stability under a dual 
A100 (80GB) GPU setup.

\subsection{Curriculum Strategy}

To stabilize early optimization and improve symbolic adaptation, 
we employed a staged curriculum approach:

\begin{itemize}
    \item \textbf{Stage 1 — Textual Warm-up:} Initial training focused 
    on abstracts, introductions, and conclusions to establish 
    linguistic fluency before exposing the model to dense symbolic content.
    
    \item \textbf{Stage 2 — Symbolic Integration:} Full LaTeX bodies, 
    including theorem environments and mathematical derivations, 
    were introduced to enable structured reasoning adaptation.
    
    \item \textbf{Stage 3 — Mixed Curriculum:} A balanced mixture of 
    prose and formula-heavy content ensured robustness across 
    explanatory and symbolic regimes.
\end{itemize}

Although the architecture supports a 4096-token context window, 
training sequences were constructed at 768 tokens to maximize 
batch throughput and maintain stable memory utilization.

\subsection{Training Time and Scaling Considerations}

Training was conducted on 2$\times$NVIDIA A100 (80GB) GPUs 
using ZeRO Stage 2 optimization and bfloat16 precision. 
The primary pretraining phase required approximately 
5{,}000--8{,}000 GPU-hours.

For a 1.36B-parameter model, the Chinchilla scaling law 
suggests an optimal training token budget of approximately 
27B tokens. Our 52.18B-token pretraining corpus therefore 
places the model in a data-rich regime ($\approx 38$ tokens per parameter), 
prioritizing domain robustness over strict compute optimality.

This configuration reflects a deliberate tradeoff: maximizing 
scientific coverage and symbolic stability under moderate 
hardware constraints rather than scaling parameter count alone.

\subsection{Optimization Details}

Training was conducted using:

\begin{itemize}
    \item AdamW optimizer with weight decay
    \item bfloat16 mixed precision
    \item ZeRO Stage 2 memory optimization
    \item Gradient checkpointing for activation memory reduction
\end{itemize}

The effective global batch size was scaled via gradient accumulation 
to maintain stable gradient statistics while fitting within GPU memory limits.

\section{Training Dynamics and Optimization Analysis}

A total of 24 experimental runs were conducted to refine 
hyperparameters, stabilize optimization, and improve hardware 
utilization. These runs varied in dataset scale, learning rate, 
gradient accumulation, and preprocessing configurations.

\subsection{Iterative Optimization Across 24 Runs}

Early experimental runs were intentionally exploratory and 
frequently unstable. Several runs terminated prematurely due 
to suboptimal hyperparameters or memory constraints. 

Notably:

\begin{itemize}
    \item \textbf{Run 24:} Trained on a reduced 20GB subset 
    to validate pipeline stability.
    \item \textbf{Run 23 and Run 20:} Trained on the full 200GB 
    processed corpus.
\end{itemize}

This progression enabled controlled scaling from small-data 
debugging to full-scale training.

\subsection{Small-Data Regime (20GB)}

\begin{figure}[H]
\centering
\includegraphics[width=0.8\textwidth]{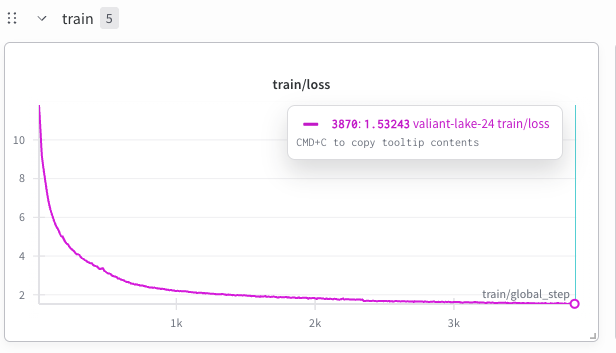}
\caption{Training loss for Run 24 (20GB dataset). Loss oscillates 
and converges slowly, indicating limited data regime.}
\label{fig:small_data_loss}
\end{figure}

Run 24, trained on approximately 20GB of data, exhibits 
unstable convergence behavior. Training loss decreases 
initially but oscillates and plateaus at a relatively 
high value (Figure~\ref{fig:small_data_loss}).

\textbf{Observation:}  
Small-scale pretraining leads to noisy gradient dynamics 
and reduced convergence efficiency. The model begins to 
memorize patterns without sufficient diversity for stable 
generalization.

\subsection{Full-Data Regime (200GB)}

Run 23 and Run 20 were trained on the full 200GB processed corpus.

\begin{figure}[H]
\centering
\includegraphics[width=0.8\textwidth]{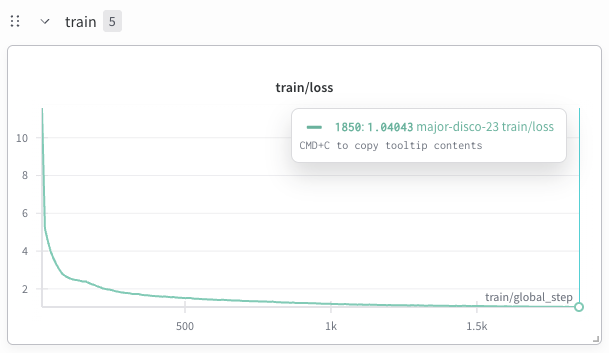}
\caption{Training loss for Run 23 (200GB dataset). Loss decreases 
smoothly with improved stability compared to small-data regime.}
\label{fig:mid_data_loss}
\end{figure}

\begin{figure}[H]
\centering
\includegraphics[width=0.8\textwidth]{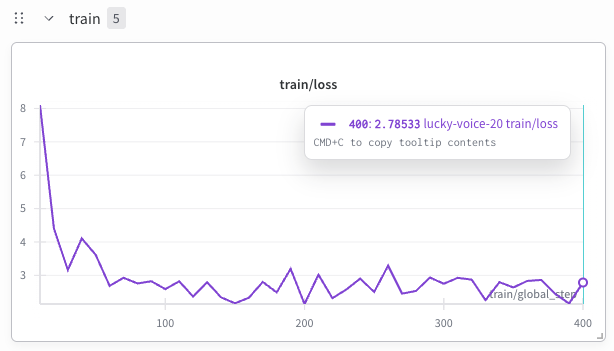}
\caption{Final optimized run (Run 20, 200GB). Training loss 
exhibits smooth monotonic convergence with long-tail stabilization.}
\label{fig:final_run_loss}
\end{figure}

\textbf{Observations:}

\begin{itemize}
    \item Loss reduction is significantly smoother under full data scale.
    \item Gradient noise is reduced relative to the 20GB regime.
    \item Convergence exhibits classic transformer long-tail behavior.
\end{itemize}

\subsection{Validation and Overfitting Analysis}

Validation loss decreased monotonically throughout training 
and remained closely aligned with training loss. Importantly, 
no sustained divergence between train and eval curves was observed.

\begin{figure}[H]
\centering
\includegraphics[width=0.8\textwidth]{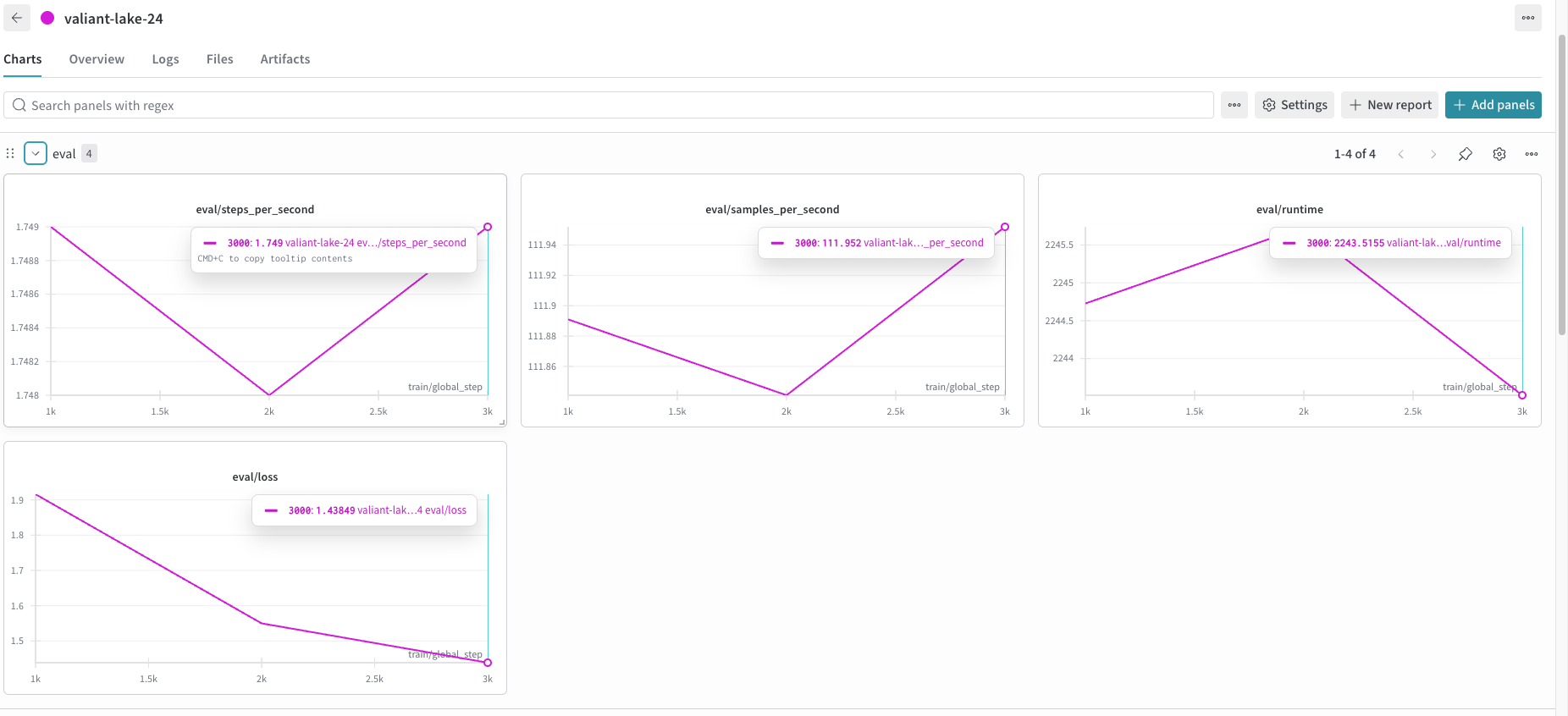}
\caption{Training and validation loss curves for the final run. 
No widening divergence is observed.}
\label{fig:train_eval}
\end{figure}

The absence of widening train–eval gaps suggests that the model 
did not enter a severe overfitting regime within the observed 
training horizon.

Final validation loss corresponds to a perplexity of approximately:

\[
\exp(1.438) \approx 4.2
\]

indicating strong adaptation to scientific corpora.

\subsection{Gradient Stability}

Gradient norm monitoring revealed:

\begin{itemize}
    \item Early warm-up spikes (expected behavior)
    \item Rapid stabilization below 1.0
    \item No late-stage explosion or vanishing gradients
\end{itemize}

\begin{figure}[H]
\centering
\includegraphics[width=0.8\textwidth]{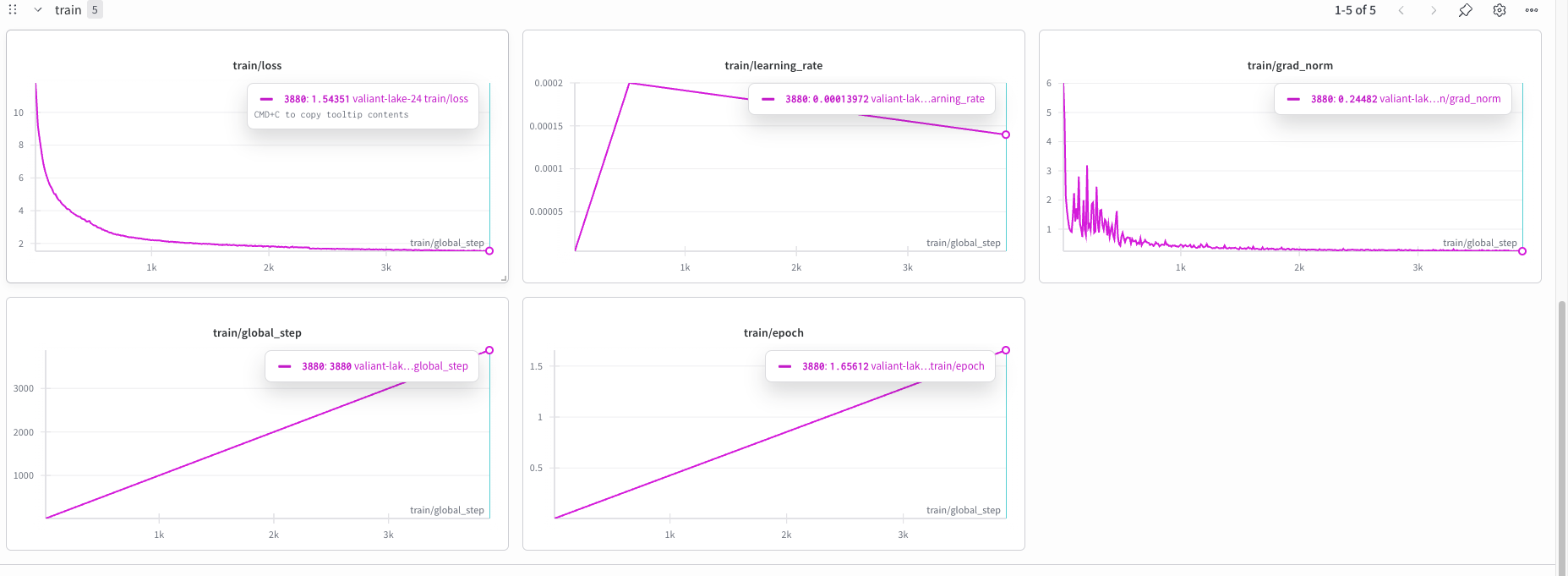}
\caption{Gradient norm across training steps. Stability is achieved 
after early warm-up.}
\label{fig:grad_norm}
\end{figure}

This confirms stable optimization under the selected 
learning rate and batch configuration.

\subsection{Hardware Utilization}

GPU monitoring indicated:

\begin{itemize}
    \item Sustained utilization above 95\%
    \item Stable power consumption (~300W)
    \item No ECC memory errors
    \item No persistent I/O stalls
\end{itemize}

\begin{figure}[H]
\centering
\includegraphics[width=0.8\textwidth]{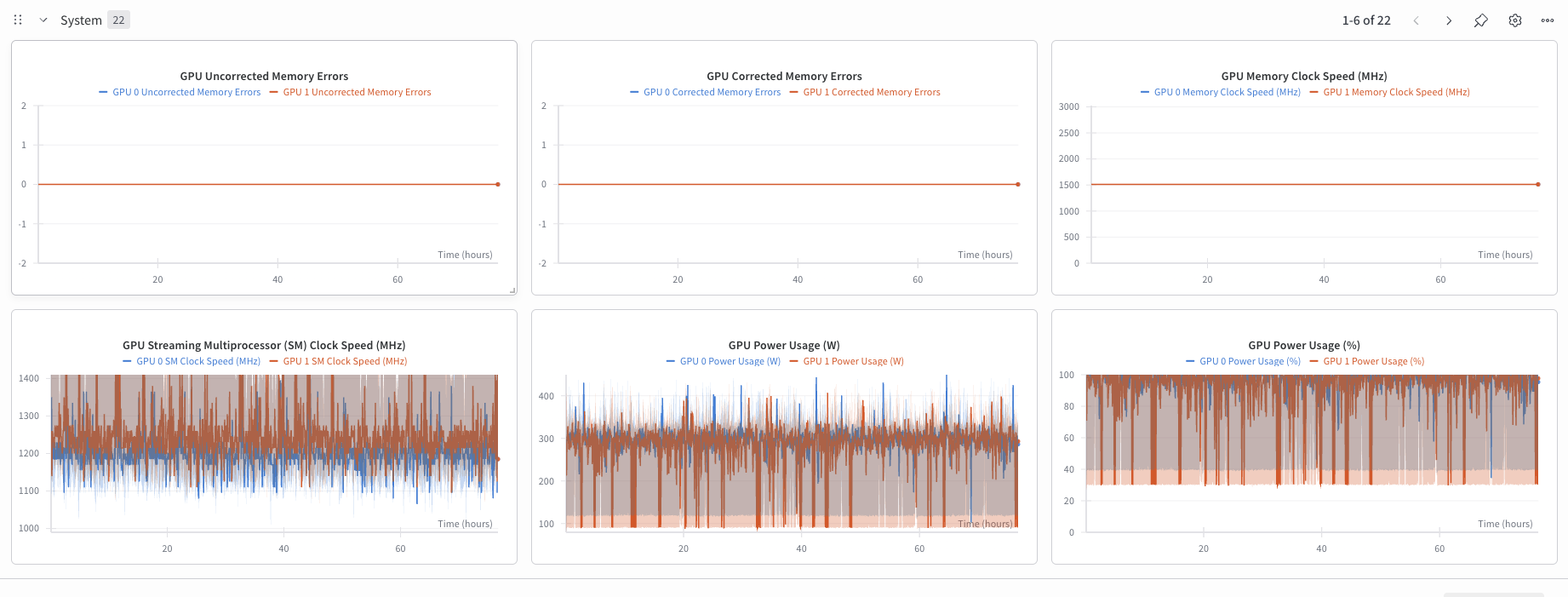}
\caption{GPU utilization and power metrics during final training run.}
\label{fig:gpu_metrics}
\end{figure}

These results indicate efficient pipeline throughput and 
effective distributed configuration.

\subsection{Best Practices Derived}

The 24-run optimization cycle yielded several practical insights:

\begin{itemize}
    \item Conservative learning rate schedules improve stability 
    in symbolic-heavy corpora.
    \item Full-scale data significantly reduces gradient noise.
    \item Monitoring gradient norms prevents silent divergence.
    \item Storage throughput can bottleneck training before compute.
    \item Iterative small-scale debugging (20GB subset) accelerates 
    stabilization prior to full-scale runs.
\end{itemize}

Collectively, these observations reinforce the importance of 
systematic experimentation and infrastructure-aware optimization 
when training small scientific language models.

\section{Evaluation}

Model evaluation was conducted primarily using perplexity 
on held-out scientific validation data.

The trained model demonstrates strong familiarity with 
mathematical notation, LaTeX structures, and formal 
scientific writing patterns. However, as the base model 
was trained exclusively on raw scientific corpora, it does 
not exhibit instruction-following or conversational behavior.

This work does not aim to compete with large-scale 
instruction-tuned systems. Instead, the evaluation focuses 
on analyzing the capabilities and limitations of a 
small, domain-specialized language model trained from 
structured scientific data.

\section{Empirical Observations}

Several practical insights emerged during training:

\begin{itemize}
    \item \textbf{Data Yield is Pipeline-Dependent:} Effective dataset 
    size was driven more by preprocessing decisions than by raw data 
    availability. Archive validation, LaTeX cleaning, and filtering 
    heuristics significantly impacted usable token volume.
    
    \item \textbf{Storage as a Bottleneck:} In early stages, I/O 
    throughput and storage constraints proved more limiting than 
    raw compute capacity.
    
    \item \textbf{Language Filtering Sensitivity:} Applying language 
    detection heuristics too early in the pipeline resulted in the 
    removal of valid scientific documents due to dense symbolic content.
    
    \item \textbf{Instruction Following Does Not Emerge Naturally:} 
    Pretraining on raw scientific corpora does not produce 
    conversational or instruction-following capabilities without 
    explicit post-training alignment.
\end{itemize}

These findings highlight the central role of data engineering 
and pipeline design in small-scale language model training, 
often outweighing architectural modifications in practical impact.

\section{Limitations and Lessons Learned}

Despite careful design and systematic engineering, several 
limitations remain.

\textbf{Compute Constraints.}  
Training was performed on a dual A100 (80GB) setup. While 
sufficient for a 1.36B-parameter model, this restricts 
exploration of larger architectures, extended context training, 
or aggressive hyperparameter sweeps. The total compute cost 
(5{,}000--8{,}000 GPU-hours) highlights the non-trivial resource 
requirements even for mid-scale language models.

\textbf{Storage and I/O Bottlenecks.}  
Raw arXiv archives, intermediate extraction artifacts, and 
processed JSONL corpora require substantial disk capacity 
and high-throughput I/O. In early stages, archive handling 
and storage bandwidth became more limiting than GPU compute.

\textbf{Preprocessing Sensitivity.}  
LaTeX extraction, archive validation, and metadata filtering 
significantly influenced final token yield. Small heuristic 
changes led to large variations in usable data volume. 
This introduces an unavoidable degree of pipeline-induced bias.

\textbf{Scaling Regime Tradeoffs.}  
Although trained on 52.18B tokens, the model contains 1.36B parameters, 
placing it in a data-heavy regime relative to compute-optimal scaling. 
While beneficial for domain specialization, this may reduce 
marginal efficiency gains compared to larger parameter models 
trained under strictly optimal token-to-parameter ratios.

\textbf{Context-Length Utilization.}  
The architecture supports a 4096-token context window; however, 
training sequences were limited to 768 tokens to maximize 
batch throughput. As a result, long-context reasoning capacity 
may not be fully realized.

\textbf{Evaluation Scope.}  
Evaluation primarily relies on perplexity over held-out 
scientific corpora. This does not directly measure reasoning 
correctness, theorem validity, or symbolic proof consistency. 
More structured mathematical benchmarks would provide 
stronger assessment of formal reasoning capabilities.

\textbf{Domain Bias.}  
The dataset is restricted to selected scientific categories 
(math, theoretical physics, and statistical learning). 
While this strengthens specialization, it limits 
general-domain adaptability.

\textbf{Lack of Instruction Alignment.}  
The base model is not instruction-tuned and therefore unsuitable 
for direct conversational deployment without additional alignment.

\textbf{Reproducibility Constraints.}  
Large-scale LaTeX preprocessing pipelines, storage requirements, 
and GPU resource needs may limit exact reproducibility for 
researchers without similar infrastructure.

Collectively, these limitations emphasize that successful 
small-scale language model training depends as much on 
infrastructure planning and data engineering rigor as on 
architectural design.

\section{Conclusion}

This paper presents an end-to-end case study of training a 
1.36B-parameter scientific language model from raw arXiv 
LaTeX sources. Beyond architectural configuration, we emphasize 
the critical role of data filtering, tokenization design, 
and curriculum strategy in enabling stable domain specialization.

Our findings suggest that for small-to-mid-scale models, 
careful data engineering can rival architectural scaling in impact. 
Moreover, transparent reporting of preprocessing decisions, 
tokenization strategies, and compute tradeoffs remains essential 
for reproducibility in language model research.

We hope this work contributes to a more realistic and 
engineering-aware perspective on scientific language modeling, 
particularly for researchers operating under constrained compute budgets.

Future work includes exploring longer-context training, 
instruction-aligned post-training strategies, and systematic 
evaluation on formal mathematical reasoning benchmarks.

\section*{Acknowledgements}
The author would like to thank the arXiv project for maintaining the open-access scientific archive used in this work. We also express our gratitude to the creators of the \textit{Proof-Pile-2}, \textit{OpenWebMath}, and \textit{MathInstruct} datasets for their contributions to the open-science community. This work was supported by computational resources utilizing NVIDIA A100 GPUs.


\newpage
\appendix
\section{Code Repository}
The code used for experiments and analysis in this work is publicly available at:
\url{https://github.com/kitefishai/KiteFish-A1-1.5B-Math}

\section{Difference between Tokenization, Chunking, Vector and Embedding}

\begin{description}
  \item Tokenization
  Split text into small units called tokens (words or sub-words).\\
  \emph{Example:} ``I love AI'' $\rightarrow$ tokens

  \item[Chunking]
  Group tokens into larger pieces so long text can be processed.\\
  \emph{Example:} Paragraph $\rightarrow$ multiple chunks

  \item[Vector]
  A numerical list that represents something.\\
  \emph{Example:} $[0.12, -0.87, 3.45]$

  \item[Embedding]
  A vector that captures meaning of a chunk (sentence, image, etc.).\\
  Similar meaning $\rightarrow$ similar vectors

  \item[Pipeline]
  Text $\rightarrow$ tokenization $\rightarrow$ chunking $\rightarrow$ model $\rightarrow$ embeddings
\end{description}

\section{Importance of \texttt{model\_type} in config.json}

In a \texttt{config.json}, \texttt{model\_type} is \textbf{not merely documentation}. It is used to:

\begin{enumerate}
  \item \textbf{Select the correct configuration class}
    \begin{itemize}
      \item \texttt{"model\_type": "llama"} $\rightarrow$ \texttt{LlamaConfig}
      \item \texttt{"model\_type": "deepseek"} $\rightarrow$ \texttt{DeepseekConfig}
    \end{itemize}

  \item \textbf{Dispatch the correct model and tokenizer classes}
    \begin{itemize}
      \item \texttt{AutoModel.from\_pretrained()}
      \item \texttt{AutoTokenizer.from\_pretrained()}
    \end{itemize}
\end{enumerate}

Internally, Hugging Face maintains mappings such as:
\begin{verbatim}
MODEL_TYPE_TO_CONFIG = {
    "llama": LlamaConfig,
    "deepseek": DeepseekConfig,
}
\end{verbatim}
Thus, \texttt{model\_type} determines \textbf{how the configuration is interpreted}.

\section{Relationship Between \texttt{model\_type} and the Tokenizer}

An important distinction is the following:

\begin{quote}
\textbf{Tokenizers are not inferred from model weights.}  
They are inferred from the \textbf{configuration} and the \textbf{tokenizer files}.
\end{quote}

\subsection{What \texttt{model\_type} Influences for Tokenization}

\begin{table}[h]
\centering
\begin{tabular}{p{4cm} c}
\hline
\textbf{Aspect} & \textbf{Influenced} \\
\hline
Tokenizer class (e.g., LLaMATokenizer) & Yes \\
Tokenizer vocabulary & No \\
BPE vs.\ SentencePiece & Yes \\
Special tokens (BOS, EOS, \texttt{<s>}, \texttt{</s>}) & Yes \\
Chat templates & Often \\
\hline
\end{tabular}
\caption{Effect of \texttt{model\_type} on tokenizer behavior}
\end{table}

\section{What Happens When Loading a Tokenizer}

When executing:

\begin{verbatim}
AutoTokenizer.from_pretrained(path)
\end{verbatim}

Hugging Face performs the following steps:

\begin{enumerate}
  \item Load \texttt{config.json}
  \item Read \texttt{model\_type}
  \item Select a tokenizer class
  \item Load tokenizer files, such as:
    \begin{itemize}
      \item \texttt{tokenizer.json}
      \item \texttt{tokenizer.model} (SentencePiece)
      \item \texttt{vocab.json}, \texttt{merges.txt}, etc.
    \end{itemize}
\end{enumerate}

\noindent
\textbf{Warning:} If the tokenizer files do not match the selected tokenizer class, the tokenizer may fail or behave incorrectly.

\section{Example: \texttt{model\_type = "llama"} with DeepSeek Architecture}

\subsection{Case A: DeepSeek Model with \texttt{model\_type = "llama"}}

This configuration is dangerous but subtle.

\subsubsection{Tokenizer Effects}

\begin{itemize}
  \item Hugging Face loads the \textbf{LLaMA tokenizer}
  \item Uses:
    \begin{itemize}
      \item SentencePiece
      \item LLaMA special tokens
      \item LLaMA BOS/EOS behavior
    \end{itemize}
\end{itemize}

If the model was trained using a \textbf{DeepSeek tokenizer}:

\begin{itemize}
  \item Token IDs no longer align
  \item Embedding rows correspond to incorrect tokens
  \item Generated output becomes meaningless
\end{itemize}

\noindent
This failure mode is typically \textbf{silent}.

\subsubsection{Model Architecture Effects}

\begin{itemize}
  \item \texttt{AutoModel} attempts to instantiate a \textbf{LLaMA architecture}
  \item DeepSeek may differ in:
    \begin{itemize}
      \item Attention implementation
      \item Rotary embedding scaling
      \item Mixture-of-Experts layers
      \item RMSNorm usage
    \end{itemize}
\end{itemize}

Possible outcomes include:

\begin{itemize}
  \item Tensor shape mismatches (hard failure)
  \item Successful loading with incorrect behavior
\end{itemize}

\subsection{Case B: DeepSeek Weights with LLaMA Tokenizer}

This configuration leads to the following issues:

\begin{table}[h]
\centering
\begin{tabular}{l l}
\hline
\textbf{Problem} & \textbf{Result} \\
\hline
Token ID mismatch & Nonsensical generation \\
Special token differences & BOS/EOS errors \\
Different vocabulary size & Runtime failures \\
Different token segmentation & Severe perplexity increase \\
\hline
\end{tabular}
\caption{Effects of tokenizer mismatch}
\end{table}

Even a single token offset is sufficient to invalidate the model.

\section{Why Some Models Appear to Work Despite Mismatch}

In some cases:

\begin{itemize}
  \item DeepSeek forks the LLaMA tokenizer
  \item Vocabulary is identical
  \item Special tokens remain compatible
\end{itemize}

In such scenarios:

\begin{itemize}
  \item Mislabeling \texttt{model\_type} may appear to work
  \item However, it breaks downstream tooling, including:
    \begin{itemize}
      \item Chat templates
      \item LoRA adapters
      \item Quantization configurations
      \item Future library updates
    \end{itemize}
\end{itemize}

This setup is fragile and unsafe.

\section{Correct Handling of DeepSeek vs.\ LLaMA}

\subsection{Recommended Configuration}

\begin{verbatim}
{
  "model_type": "deepseek",
  "architectures": ["DeepseekForCausalLM"],
  "vocab_size": 102400
}
\end{verbatim}

The tokenizer files must match \textbf{exactly} those used during training.

If DeepSeek intentionally uses the LLaMA tokenizer:

\begin{itemize}
  \item Keep tokenizer files identical
  \item Still use \texttt{model\_type = "deepseek"}
\end{itemize}

\end{document}